\documentclass[sigconf, authorversion,nonacm]{acmart}

\usepackage{xspace}
\usepackage{multirow}
\usepackage{balance}
\usepackage{enumitem}
\usepackage{hhline}
\usepackage[noend,ruled]{algorithm2e}
\usepackage{array}

\newcommand{\ie}{\textit{i.e.}, }
\newcommand{\eg}{\textit{e.g.}, }

\AtBeginDocument{%
  \providecommand\BibTeX{{%
    \normalfont B\kern-0.5em{\scshape i\kern-0.25em b}\kern-0.8em\TeX}}}

\begin{document}

\title{PetalView: Fine-grained Location and Orientation Extraction of Street-view Images via Cross-view Local Search}

\author{Wenmiao Hu}
\affiliation{%
  \institution{Grab-NUS AI Lab, \\ National University of Singapore}
  \country{}
}
\email{hu.wenmiao@u.nus.edu}

\author{Yichen Zhang}
\affiliation{%
  \institution{Grab-NUS AI Lab, \\ National University of Singapore}
  \country{}}
\email{zhang.yichen@u.nus.edu}

\author{Yuxuan Liang}
\authornote{Work was done while at National University of Singapore.}
\affiliation{%
  \institution{The Hong Kong University of Science and Technology (Guangzhou)}
  \country{}
  }
\email{yuxliang@outlook.com}

\author{Xianjing Han}
\affiliation{%
  \institution{Grab-NUS AI Lab, \\ National University of Singapore}
  \country{}
  }
\email{xianjing@nus.edu.sg}

\author{Yifang Yin}
\affiliation{%
  \institution{Institute for Infocomm Research, A*STAR}
  \country{}}
\email{yin_yifang@i2r.a-star.edu.sg}

\author{Hannes Kruppa}
\affiliation{%
  \institution{Grabtaxi Holdings Pte. Ltd.}
  \country{}}
\email{hannes.kruppa@grabtaxi.com}

\author{See-Kiong Ng}
\affiliation{%
  \institution{Grab-NUS AI Lab \& \\ Institute of Data Science,\\ National University of Singapore}
  \country{}}
\email{seekiong@nus.edu.sg}

\author{Roger Zimmermann}
\affiliation{%
  \institution{Grab-NUS AI Lab, \\ National University of Singapore}
  \country{}}
\email{rogerz@comp.nus.edu.sg}

\renewcommand{\shortauthors}{Wenmiao Hu et al.}

\begin{abstract}

Satellite-based street-view information extraction by cross-view matching refers to a task that extracts the location and orientation information of a given street-view image query by using one or multiple geo-referenced satellite images. Recent work has initiated a new research direction to find accurate information within a local area covered by one satellite image centered at a location prior (\eg from GPS). It can be used as a standalone solution or complementary step following a large-scale search with multiple satellite candidates. However, these existing works require an accurate initial orientation (angle) prior (\eg from IMU) and/or do not efficiently search through all possible poses. To allow efficient search and to give accurate prediction regardless of the existence or the accuracy of the angle prior, we present PetalView extractors with multi-scale search. The PetalView extractors give semantically meaningful features that are equivalent across two drastically different views, and the multi-scale search strategy efficiently inspects the satellite image from coarse to fine granularity to provide sub-meter and sub-degree precision extraction. Moreover, when an angle prior is given, we propose a learnable prior angle mixer to utilize this information. Our method obtains the best performance on the VIGOR dataset and successfully improves the performance on KITTI dataset test~1 set with the recall within 1 meter (r@1m) for location estimation to 68.88\% and recall within 1 degree (r@1$^\circ$) 21.10\% when no angle prior is available, and with angle prior achieves stable estimations at r@1m and r@1$^\circ$ above 70\% and 21\%, up to a 40$^\circ$ noise level.

\end{abstract}

%
%




\maketitle
\begin{figure}
  \includegraphics[width=0.48\textwidth,trim = 2.6cm 3.8cm 8cm 1.1cm, clip]{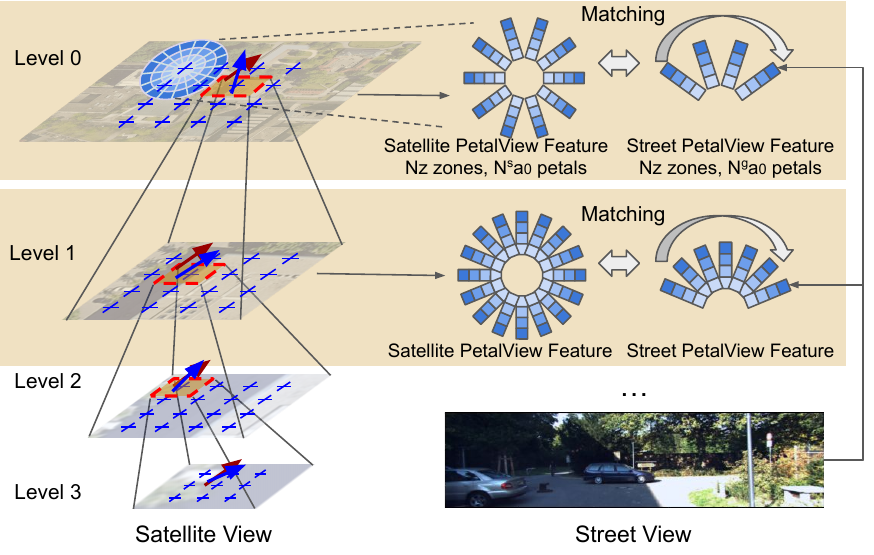}
  \vspace{-3mm}
  \caption{PetalView in multi-scale search. Blue arrow: predicted direction \& location, brown arrow: ground-truth, blue cross: search anchors. PetalView features are extracted through all levels from the two views, and they are matched to find the optimal location and orientation.}
  \label{fig:multi-level-petal}
  \vspace{-3mm}
\end{figure}
\section{Introduction}
Street-view images are used in many downstream computer vision tasks to extract information in a wide range of applications, \eg for autonomous driving~\cite{JiangEtAl2022, LiEtAl2022a, MaEtAl2022a}, urban management~\cite{ManiatEtAl2021, WangVermeulen2021, WangEtAl2021, GMEtAl2021}, map making~\cite{SahaEtAl2022, BalaliEtAl2015}, trajectory tracking~\cite{RegmiShah2021}. However, not all street-view images are collected with accurate location/orientation information due to 1) a lack of additional devices (\eg GPS, IMU) and 2) acquisition errors caused by poor sensor estimations. Hence, finding/improving accurate location and orientation information of street-view images becomes a critical problem. 

To solve this issue, satellite imagery is used as an alternative source to extract coarse/fine-grained location/orientation information of the street-view images via cross-view matching. Existing studies~\cite{ShiEtAl2020, ShiEtAl2019a, ZhuEtAl2022, ZhangEtAl2022, HuEtAl2022a, HuEtAl2018, WorkmanEtAl2015} work on the large-scale search to find the best-matched satellite candidate from a large pool of satellite images for a query street-view image. The center location of the satellite image is recognized as the location of the street-view image, which is referenced as the image retrieval-based geo-localization. On top of this, a few papers~\cite{HuEtAl2022a, ZhuEtAl2021a} work on fine-grained orientation extraction or sub-image location extraction when the orientation of the street-view image is known. However, it is still difficult to solve both challenges in \emph{large-scale search} effectively and simultaneously due to the complexity of the problem and enormous 3D search space (2D for location + 1D for azimuth orientation).

The reasons for investigating fine-grained orientation and location prediction in \emph{local-scale search} within one reference satellite image are twofold: 1) it is a standalone solution to refine the location and orientation information when a prior location in other modalities is given (\eg GPS, text tags), and 2) it can be used as a complementary step following a large-scale search. Recent work LM~\cite{ShiLi2022} initiates the research in this direction to find the fine-grained location of the street-view image within one reference satellite image and refines the orientation when the angle noise level is within a small perturbation range.
The proposed method, which leverages the Levenberg-Marquardt (LM) algorithm, requires a reasonable initial guess, \ie it only works well when the prior angle is available and not too far from the ground truth. Most of the time having such accurate orientation prior
is not feasible. Hence, we aim to provide a generic solution without using an angle prior.

Inspired by prior work, our study extends this direction with the goal to 1) expand the use cases to allow prediction without angle prior and provide stable performance regardless of the existence or accuracy of the prior angle information, 2) broaden the search space and improve the extraction precision with limited feature queries to provide efficient search and finer-grained results.  

To achieve this goal, we propose a novel PetalView representation that structures and limits the accessible information of queries from street-view and satellite search anchors, and a dynamic multi-scale search strategy to improve the result precision to sub-meter and sub-degree levels with limited queries. \autoref{fig:multi-level-petal} shows the conceptual illustration of the PetalView representation and multi-scale search strategy. To boost the finer-grained search, we propose to extract the PetalView feature, comprising multiple petals, which are defined based on viewing angles. Furthermore, each petal contains information dissected into different zones by distance. Each anchor in the satellite view has limited access to pixels of the image feature map based on the relative distance and angle direction. Paired with the multi-scale search strategy, which uses the best-matched anchor as the search center for the next level search and dynamically deploys finer grids. We enable pixel-level search precision in the feature space without an exhaustive search at every pixel in the extracted feature maps. Our method achieves higher and more stable overall performance through different use cases. The main contributions of the work lie in the following aspects:
\begin{itemize}[leftmargin=*]
    \item Propose a PetalView network for extracting semantic-meaningful PetalView features from street and satellite views that are equivalent and can be directly used for cross-view matching.
    \item Develop a multi-scale search strategy to dynamically refine the search area from coarse to fine, and an associated batch-wise early stopping training method.
    \item Present a learnable prior angle mixer, which learns how to fuse the angle prior in other modalities into the similarity-based origination estimation curve to guide the final outcome.
    \item Conduct extensive experiments and ablation studies. Our method improves the location and orientation estimation to 68.88\% on r@1m and 21.10\% on r@1$^\circ$ without angle prior, and gives stable estimation at least 70\% and 21\% for r@1m and r@1$^\circ$ when the angle prior contains noise up to 40$^\circ$.
\end{itemize}

\section{Related Work}
\textbf{Large-scale cross-view search.} Large-scale cross-view matching refers to the process which finds the location and/or orientation of a query street-view image by identifying the satellite image candidate with the highest similarity. The input query can be a 360$^\circ$ panorama or have a limited field of view (FOV) and the extraction condition and outcomes can be the following: a) image-based location retrieval with known camera pose orientation that gives the center of the matched satellite image as the extraction result~\cite{LiuLi2019, RegmiShah2019, ShiEtAl2019a, RodriguesTani2020, WangEtAl2021a, ZhangEtAl2022, TokerEtAl2021}, b) image-based location retrieval with unknown camera orientation~\cite{VoHays2017, HuEtAl2018, CaiEtAl2019, YangEtAl2021b, ZhuEtAl2022}, c) image-based location retrieval and orientation extraction of the query image~\cite{VoHays2017, CaiEtAl2019, ShiEtAl2020, ZhuEtAl2020, HuEtAl2022a}, d) fine-grained location extraction at sub-image level with known orientation~\cite{ZhuEtAl2021a}. However, as satellite images have a large ground coverage, when the street-view images have unknown orientations, searching through every possible pixel location and direction for fine-grained extraction becomes extremely time-consuming and unpractical.

\noindent\textbf{Local-scale cross-view search.} To reduce the complexity of the problem, we can limit the search space with prior knowledge of location and orientation, typically from GPS and IMU readings. 
Recent work LM~\cite{ShiLi2022} for the first time proposes the searching of fine-grained orientation and location at degree and meter levels for street-view image queries within only one satellite image. However, it uses the Levenberg-Marquardt (LM) algorithm to gradually find the correct camera pose and wrap the satellite view feature to street view. Its nature requires a low initial noise level and cannot handle cases where the angle prior is missing or have a large error.
MCC~\cite{XiaEtAl2022} works in this domain but focuses more on the localization, the orientation is extracted by sending multiple rotated copies of street-view image, which gives limited granularity and efficiency. A concurrent work \cite{LentschEtAl2022} proposes SliceMatch, which restricts the slice-based feature by viewing angles. However, it has a few drawbacks: 1) the restriction is only on viewing angles without range/zone limits, which results in descriptors that contain different amounts of information and may create unfair comparisons for similarity-based optimization. 2) It uses a regular flat search grid that brings a trade-off between the computational resources and result precision. 3) The method requires information flow from the query image to the satellite view, which means the method can only process ad hoc. This limits the possibilities of pre-processing and usage in large-scale search.

\noindent\textbf{Birds-Eye-View transformation.} Another related topic is the transformation from street-view into Birds-Eye-View (BEV) for different purposes~\cite{LiuEtAl2022, LiEtAl2022a, JiangEtAl2022, GongEtAl2022a, PengEtAl2022, SahaEtAl2022}. Although BEV transformation focuses on one direction from street-view to BEV, these related studies provide transferable knowledge about the geometric relationship between the two views. 
Saha et al.~\cite{SahaEtAl2022} proposed a polar ray-based process to extract the BEV road map. The main motivation is to leverage the assumption that each ray passing through the camera location in an overhead map has a 1-to-1 corresponding vertical scan line in the street-view image. In this work, we use this concept to extract semantic-meaningful PetalView features that are equivalent across views to bridge the gap between the two drastically different views.
\section{Approaches}

\begin{figure}
  \includegraphics[width=0.4\textwidth,trim = 2.5cm 4cm 8cm 2.1cm, clip]{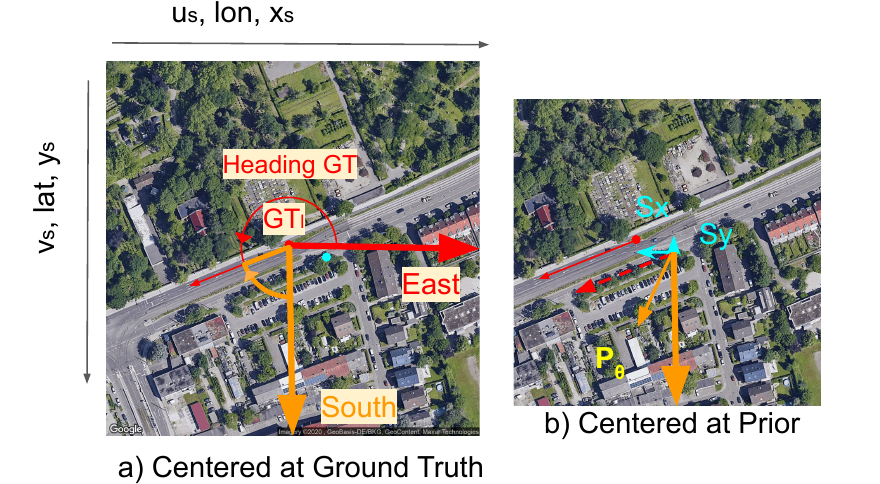}
  \vspace{-3mm}
  \caption{Earth-centric setting and ground truth generation.}
  \label{fig:settings}
  \vspace{-3mm}
\end{figure}
\begin{figure*}
  \includegraphics[width=0.8\textwidth,trim = 0cm 3.2cm 3.2cm 0cm, clip]{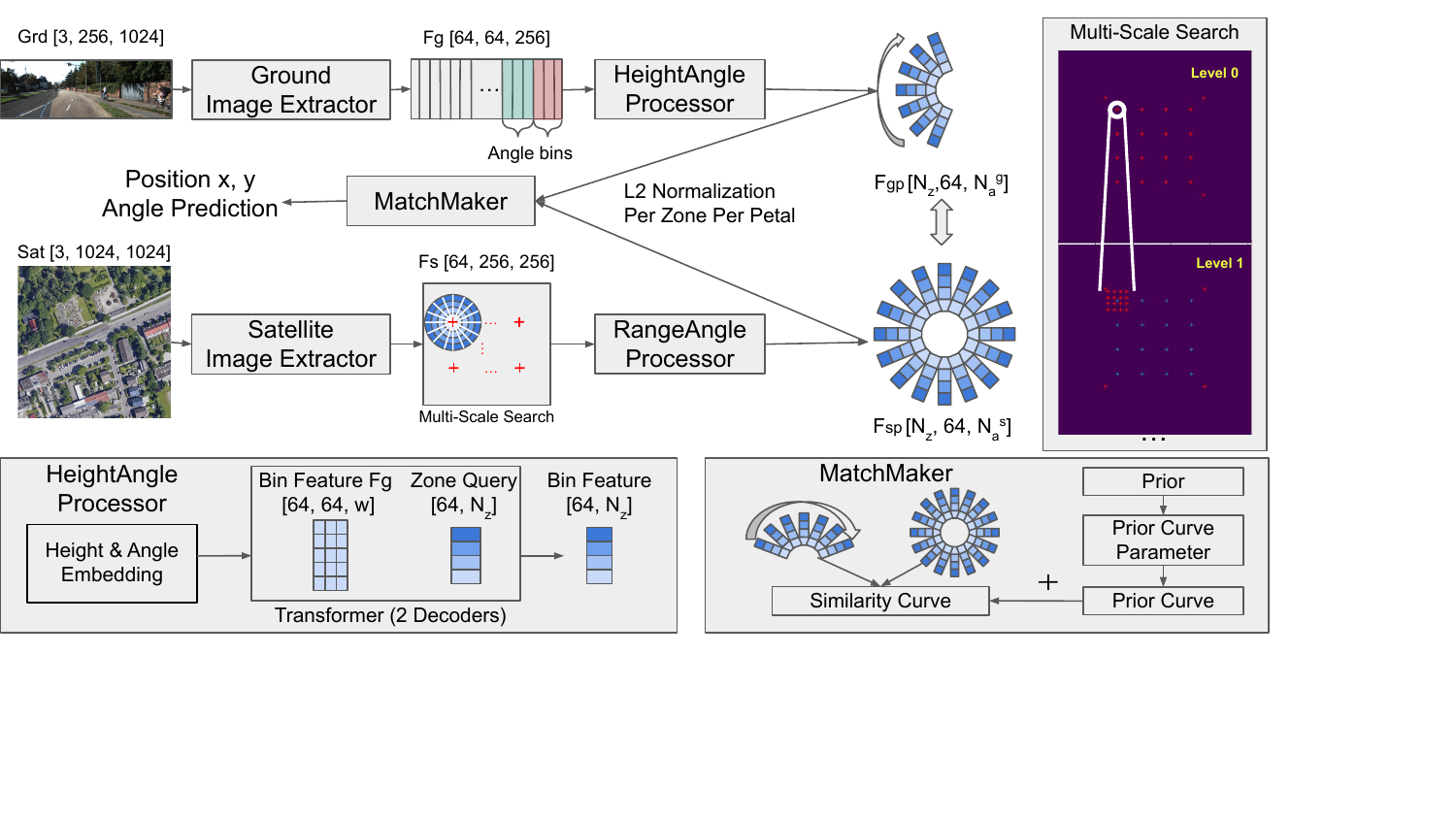}
  \vspace{-3mm}
  \caption{Overall network architecture for PetalView.}
  \label{fig:network}
  \vspace{-3mm}
\end{figure*}

\subsection{Problem Setting} \label{section: problem setting}
For local search, we define the search area within one satellite image, assuming prior location information $P_l(L_x, L_y)$ is given (the cyan dot in \autoref{fig:settings}a). This location prior $P_l$ is usually not accurate and is away from the ground truth location $\text{GT}_l(\text{GT}_x, \text{GT}_y)$ by a location shift $(S_x, S_y)$. There may exist an angle prior $P_\theta$ in the south-aligned angle error system, which is represented by the orange arrow aligning the left edge of the street-view image and the south direction. This angle prior $P_\theta$ contains estimation errors and is away from the ground truth orientation $\text{GT}_\theta$. Given $P_l$ with/without $P_\theta$, we aim to recover $\text{GT}_l$ and $\text{GT}_\theta$ at the same time. 

As for the dataset, $\text{GT}_l$ and $\text{GT}_\theta$ are given. For testing and training, the satellite image is cropped at the center at the prior $P_l$. An angle prior $P_\theta$ is an optional input, usually accompanied by a noise range $\Delta_\theta$. Previous work \cite{ShiLi2022} uses a car-centric setting, which rotates the satellite images to align the car heading direction to the east direction. We prefer an Earth-centric setting, which does not rotate the satellite images, for the following reasons. a) It allows the use cases without angle priors. b) In a car-centric setting, the location errors are evaluated based on the camera pose, which is unknown and cannot be fully recovered. c) An Earth-centric setting is more generic and can be directly used for downstream tasks.

\subsection{Overview of Our Method}

\noindent\textbf{PetalView feature extraction.} 
We propose the PetalView representation that provides semantically-meaningful features, giving equivalent information across different views. The PetalView representation dissects the information of an observation point on the image by relative angles (petals) and distance (zones), thereby limiting what is accessible by each observation. \autoref{fig:network} shows the conceptual examples of the features extracted from the two views.

Images from the two views are first processed by ResUnet-based~\cite{HeEtAl2015, RonnebergerEtAl2015, LinEtAl2016} image extractors. The resulting image features are sampled into different sub-groups and then processed to form PetalView features. At different levels, one PetalView feature $F_{gp}$ is extracted from the street-view image, while for each anchor in the satellite reference image one PetalView feature $F_{sp}$ is extracted, forming a set of $N_\text{anchor}$ PetalView features. The details of sampling and processing are explained in Section~\ref{section:petal}.

\noindent\textbf{Multi-scale search.} We propose a multi-scale search strategy that greatly reduces the number of queries on the satellite image feature map. The top-right of \autoref{fig:network} shows the zoom-in from level 0 to level~1. At each search level, the search area is regularly divided into patches and an anchor is assigned to each patch. For every level, the optimal anchor and corresponding patch are used as the new search center and search area for the next level. New anchors are dynamically assigned until the patch size is reduced to one pixel in the feature space, and the last level only searches in the closeby neighborhood. To achieve pixel-level precision, given an initial search area with size $[N \times N]$ and a grid size $N_s \times N_s$, the number of queries on the satellite image is reduced from $\mathcal{O}(N^2)$ to less than $\mathcal{O}(N_s^2 \lfloor\log_{N_s} (N)+1\rfloor)$. To complement this strategy in training, we define batch-wise early-stopping to dynamically control the depth of the search levels. The details are given in Section~\ref{section:multi-scale}.

\noindent\textbf{MatchMaker with prior angle mixer.} To extract the location and orientation after getting the PetalView features, the MatchMaker rotates the street-view feature over the satellite features for all anchors and finds the best-matched orientation which gives the highest similarity score for each anchor with sub-angle-bin precision.  When an angle prior $P_\theta$ is given, the MatchMaker dynamically learns a weighted prior Gaussian-based curve that fuses the prior angle information on the calculated similarity curve to guide the search for optimal orientation for each anchor. 
After extracting the optimal orientation and similarity score for each anchor $\{\phi_{N_\text{anchor}}\}$ at the given level, we found the sub-anchor position for the location extraction at that level with the highest similarity score after interpolation. The orientation of best-matched anchors is used as the extracted final orientation. The details are introduced in Section~\ref{section:prior}.

\subsection{PetalView Feature Extraction}\label{section:petal}

After obtaining the image features $F_s$ and $F_g$ from satellite and street view, we extract semantically-equivalent PetalView features from both branches. \autoref{fig:network} demonstrates the structure of PetalView feature. Each petal of the feature represents a slice of viewing angle $\theta_a$ from the observation points, which are the anchors in the satellite view and camera location for the street-view. To make representations richer, we leverage the distance to observation points to dissect the information into $N_z$ different zones for finer details. Each PetalView feature $F_p$ learns representations centered at the observation points, containing feature with a dimension of $[N_a, C, N_z]$, where $C$ is the depth of the feature map, $N_z$ is the number of zones and $N_a$ ($N_a^s$=360$^\circ/\theta_a$ or $N_a^g$ = FOV$/\theta_a$) is the number of angle bins (slices). As the images from the two views are taken from drastically different perspectives, we define the following processes to extract the semantic-equivalent PetalView features from both views.

\subsubsection{Street-view PetalView feature extractor} Following the assumption of \cite{SahaEtAl2022}, each column of a street-view image has a 1-to-1 correspondence polar ray in the satellite imagery over the camera center based on the slices of the viewing angle. Hence, to match each viewing angle span into the corresponding petals, we divide the street-view feature into sub-groups with nearby columns as a sampling pre-processing. For street-view processes, the difficulty is to get the zoning cues in the PetalView feature. Without using the depth information/estimator, we propose a transformer-based~\cite{VaswaniEtAl2017} HeightAngle processor to gather the zone-based feature for each petal. Formally, given street-view image feature $F_g \in \mathbb{R}^{C \times H \times W}$, the image feature is rearranged based on the required viewing span $F_g' \in \mathbb{R}^{N_a^g \times C \times H \times w_a}$, where $N_a^g = W / w_a$ and $w_a$ is the equivalent number of columns for the viewing span $\theta_a$. Let $Q_g \in \mathbb{R}^{C \times N_z}$ be the zone query, which are learnable parameters of HeightAngle processor that represent the different zones by distances. The PetalView feature $F_{gp} \in \mathbb{R}^{N_a^g \times C \times N_z}$ is extracted as:

\begin{equation}
\small
\begin{split}
    \text{Dec}(Q,K,V) =  \text{MLP}&(Q +\text{Attn}(Q,K,V))\\ F_g^{\text{emb}'} = \text{2D-}&\text{EMBD}(F_g', H, w_a),\\
    Q_g' = \text{Dec}(Q_g,F_g^{\text{emb}'},F_g^{\text{emb}'}), \;&
    F_{gp} = \text{Dec}(Q_g', F_g^{\text{emb}'}, F_g^{\text{emb}'}),
\end{split}
\end{equation}
where we perform 2D embedding on pixel height and relative angle distance to the center column for the street-view image feature $F_g'$ and ``Attn'' represents the multi-head attention.

\begin{figure}
  \includegraphics[width=0.4\textwidth,trim = 4.3cm 4.5cm 9.2cm 3cm, clip]{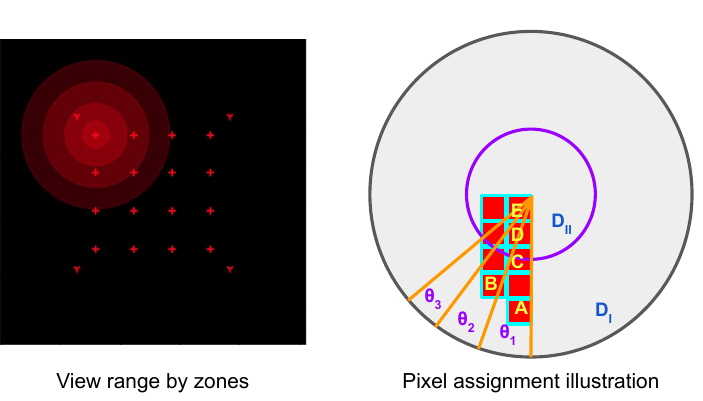}
  \vspace{-3mm}
  \caption{Petal sampling. left: viewing range for each zones. right: an illustration for pixel assignment.}
  \label{fig:petal_samples}
  \vspace{-5mm}
\end{figure}

\subsubsection{Satellite-view PetalView feature extractor} In satellite-view, the distance to the anchors and the viewing angles of each pixel can be calculated explicitly. However, as each petal and zone covers irregular shapes of regions in the satellite image and only the pixels of the correct petals and zones are used for the extraction, we need to develop efficient pre-processing to sample pixels concurrently for all viewing angles, zones, anchors and training samples of the mini-batch. Our solution is 1) to build an effective look-up table (LUT) that can sample and rearrange the pixels in the image feature according to the zones and angle specification. 2) dynamically adapt the LUT based on the location of the anchors for each training sample. 

It is a non-trivial task to build the LUT as a pixel in the image feature map may contribute to multiple viewing angles and zones, especially near the observation center. \autoref{fig:petal_samples} (right) shows an illustration. Given angle slice $\theta_1, \theta_2, \theta_3$ and range zone $D_I$ and $D_{II}$, pixel A overlaps with $(\theta_1, D_{I})$ only, but pixel B-E covers multiple angles and/or zones with ambiguous belongingness. To solve this issue and to reduce the overlaps between the pixels in each viewing angle and zone, we define the belongingness of a pixel $P_\gamma(y_\gamma, x_\gamma)$ to petal angle-range area $\text{Petal}_{ij}(\theta_i, D_{j})$ for an anchor centered at $P_c(y_0, x_0)$ by computing the angle ($\theta$) and distance (D) of the 4 corners of $P_\gamma$.
\begin{equation}
    \small
    \begin{split}
    \theta_{P_\gamma} = \{\theta_{bl}, \theta_{br}, \theta_{tl}, \theta_{tr}\}, \; D_{P_\gamma} = &\{D_{bl}, D_{br}, D_{tl}, D_{tr}\}, \\
    \text{Range}\theta_{P_\gamma} = ((\text{min}(\theta_{P_\gamma}),& \text{max}(\theta_{P_\gamma})), \\
    \text{Range}D_{P_\gamma} = ((\text{min}(D_{P_\gamma}), &\text{max}(D_{P_\gamma})),
    \end{split}
\end{equation}

\noindent We calculate the effective angle range $\Theta_{P_\gamma}$ and distance range $\mathcal{D}_{P_\gamma}$ of ${P_\gamma}$ as the overlapping region of $\text{Range}\theta_{P_\gamma}$ and $\text{Range}D_{P_\gamma}$ with the coverage of $\text{Petal}_{ij}$ $\text{Range} {\theta_i}$ and $\text{Range}D_i$. The contribution and value of the viewing angle to $\text{Petal}_{ij}$ are calculated as:
\begin{equation}
    \small
    \begin{split}
    \text{Contr}_\theta({P_\gamma}, \text{Petal}_{ij}) &= \Theta_{P_\gamma}/\text{Range} {\theta_i}, \\
    \text{Value}_\theta({P_\gamma}, \text{Petal}_{ij}) &= \Theta_{P_\gamma}/\text{Range}\theta_{P_\gamma},
    \end{split}
\end{equation}
For viewing angle selection, the pixel is accepted to $\text{Petal}_{ij}$ if:
\begin{equation}
    \small
    \Theta_{acc} (P_\gamma)=
        \begin{cases}
            1 & \text{Contr}_\theta({P_\gamma}, \text{Petal}_{ij})\! \geq \! T_1 \:
             \text{or Value}_\theta({P_\gamma}, \text{Petal}_{ij}) \! \geq \!T_1 \\
              & \text{or (Contr}_\theta({P_\gamma}, \text{Petal}_{ij}) \!<\! T_1 \! \& \!\text{ Value}_\theta({P_\gamma}, \text{Petal}_{ij})\! \geq \! T_2),\\
            0 & \text{otherwise},
        \end{cases}
\end{equation}
where $T_1=0.5$ and $T_2=0.8$. And for distance selection:
\begin{equation}
    \small
    \mathcal{D}_{acc} (P_\gamma)=
        \begin{cases}
            1 & ,\mathcal{D}_{P_\gamma} >0, \\
            0 & ,\text{otherwise},
        \end{cases}
\end{equation}

\noindent $P_\gamma$ is accepted by $\text{Petal}_{ij}$ if $\Theta_{acc} (P_\gamma)=1 \; \& \; \mathcal{D}_{acc} (P_\gamma)=1$. Based on the aforementioned selection strategy, a LUT is prepared for the $S$ number of sampled pixels at each level based on the angle slices $\theta_i, i \in [0, N_a^s)$ ($N_a^s = 360^\circ/\theta_a$) and range zones $D_i, i \: \in [0, N_z)$. Since each $\text{Petal}_{ij}$ contains different numbers of samples (\autoref{fig:level0_sample} in Appendix~\ref{section:level0 samples}), we pad $S$ to the max number of samples at current zone at each level. In training and testing, the center locations of the anchors are added to the LUT to extract the corresponding pixels $F_{ss} \in \mathbb{R}^{N_z \times N_a^s \times C\times S}$ from satellite image feature $F_s$ for each $\text{Petal}_{ij}$. For each level, $N_a^s$ and $N_z$ may be different. Hence, we refer petals across different levels $N_l$ as $\text{Petal}_{ijk}$, where $i \in [0, N_{ak}^s)$, $j \in [0, N_{zk})$ and $k \in [0, N_l)$. The visualization for the petals in scale level 0 is shown in \autoref{fig:level0_sample} in Appendix~\ref{section:level0 samples}.

After sampling the pixels for each $\text{Petal}_{ijk}$, we process the samples zone by zone with the RangeAngle processor (\autoref{fig:network}). It has a similar structure as HeightAngle Processor, which contains a learnable zone query $Q_s \in \mathbb{R}^{C \times N_z}$, only that 1) the sampled pixels within the matched zones are used for the attention; 2) instead of using the image height and the angle, RangeAngle uses the effective pixel distance to the zone center and the effective pixel angle distance to the petal center for the 2D embedding. The extracted PetalView feature for satellite view is $F_{sp} \in\mathbb{R}^{N_a^s \times  C\times N_z}$.

\subsection{Multi-Scale Search} \label{section:multi-scale}
\begin{figure}
  \includegraphics[width=0.38\textwidth,trim = 0.5cm 3cm 16cm 1cm, clip]{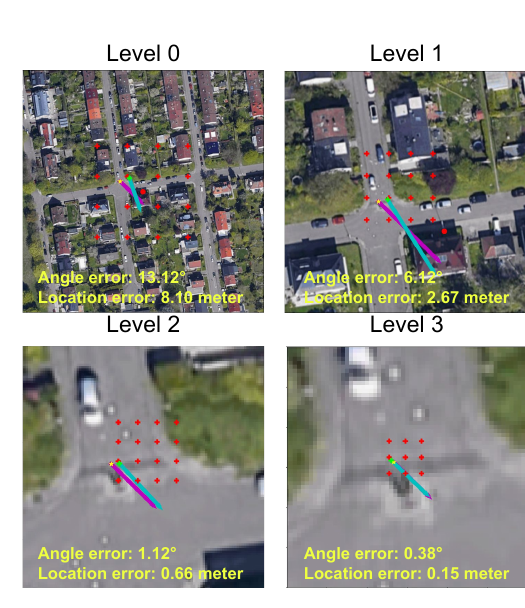}
  \vspace{-3mm}
  \caption{Anchors at each level. The selected anchors at the first three levels are [1,1], [1,2], [0,2] in columns and rows. Cyan arrow: detected direction \& location, pink arrow: ground-truth, red cross: anchors, red circle: image center.}
  \label{fig:multi-scale}
  \vspace{-3mm}
\end{figure}

In Section~\ref{section:petal}, we introduce the PetalView feature for any observation anchor. However, an exhaustive search through the entire satellite feature map is not practical. To reduce the number of queries required for obtaining pixel-level precision search in the feature map space, we propose a multi-scale search. It restructures classic flat search on a regular grid into a multi-level hierarchical search with increased granularity at each level. We take the feature map $F_s$ with a size of [$H_s$, $H_s$] as in the illustration and a $N_l$ level search strategy is introduced. Assuming the satellite imagery is usually cropped with a side length overlapping of 50\%, the search area reduces to the center half. 

We define multi-scale search following \autoref{alg:multi-scale} in Appendix~\ref{appex: multi-scale}. For each level besides the last level, $N_s \times N_s$ anchor points are distributed around the search center $X_c$. The search center is the image center at level 0 and the previous optimal anchor for later levels. When the distance between the anchors has been reduced to 1 pixel, a final search level is created with anchors $N_s' \times N_s'$ for pixel-level granularity. Combined with PetalView extractors, the granularity of the viewing-range and angles can be set differently, \eg from coarse to fine. \autoref{fig:multi-scale} shows the anchors created with the results. As for the search area of $[128, 128]$ in the feature space, using a four-level search reduces the number of queries from $16,384 = 128^2$ times to $ 57 = 3 \times 16 + 9$ times and the accuracy increases drastically at each level.

For training, we dynamically control the learning depth by monitoring the output at each level, allowing early stopping of the search and sample masking. 1) Early stopping: if none of the samples in the mini-batch find the correct anchor, the search stops and the model is updated with all accumulated loss through processed levels. 2) Sample masking: if the search continues to the next level but some of the mini-batch samples already miss the correct anchor, then the loss of these samples is no longer calculated from the next level.

\subsection{MatchMaker with Prior Angle Mixer} \label{section:prior}

After obtaining the PetalView features $F_{gp}$ and $F_{sp}(m), m \in N_\text{anchor}$, the MatchMaker calculates the optimal orientation between the $F_{gp}$ and all $F_{sp}(m)$ that gives the highest similarity. We adapted the curve-smoothing (CS($\cdot$)) method of~\cite{HuEtAl2022a} to estimate the optimal orientation at sub-petal precision with a scaling factor of 5. The optimal orientation between $F_{gp} \in \mathbb{R}^{N_a^g \times C \times N_z} $ and $F_{sp} \in \mathbb{R}^{N_a^s \times C\times N_z}$ is the angle rotation position $w_\text{opt}$ that gives the highest cross-correlation value between the two features, which is calculated as: 
\begin{equation}
\small
\label{eq:correlation}
\begin{split}
\operatorname*{argmax}_w ((F_{gp} \star F_{sp})[w]) &= \operatorname*{argmax}_w(\text{CS}(\sum\nolimits_{z=1}^{N_z} \sum\nolimits_{c=1}^{C}\sum\nolimits_{a=1}^{N_a^g} \\ &(F_{gp} [a, c, z] F_{sp} [\text{mod}(a+w, {N_a^g}) \: , c, z]))),
\end{split}
\end{equation}
where $w_\text{opt}$ can achieve sub-petal precision of one-fifth of $\theta_a$. For our case with 4-level search, the granularity reaches 0.5 degrees.

When an angle prior $P_\theta$ and the noise level $\Delta_p$ are given, an additional Gaussian-based prior curve is added on top of the cross-correlation curve. Two additional learnable parameters $\delta_p$ and $\rho_p$ are assigned to the MatchMaker to balance the impact of the prior curve to the original estimation. The prior curve and prior-based orientation estimation are given as:
\begin{equation}
\small
\label{eq:prior_angles}
\begin{split}
P(P_\theta,\Delta_p, \rho_p, \delta_p)[w] =& \frac{\rho_p}{(0.5 \delta_p \Delta_p)\sqrt{2\pi}}\exp(-\frac{\text{wrap}((w-P_\theta))^2}{2((0.5 \delta_p \Delta_p))^2}), \\
w_\text{opt} = \operatorname*{argmax}_w & ((F_{gp} \star F_{sp})[w] + P(P_\theta,\Delta_p, \rho_p, \delta_p)[w]),
\end{split}
\end{equation}

\noindent When the anchors are far away from the correct location, the similarity scores are relatively low and the prior knowledge has a higher influence, compared to when the anchor is close to the correct location (\autoref{fig:prior_curve} in Appendix~\ref{section: appendix prior}). For each search level, the similarities at the anchors are interpolated to find the sub-anchor location optimal $X_\text{opt}$. The orientation of the best-matched anchor is taken as the angle prediction result. \autoref{fig:refine_location} in Appendix~\ref{section: refinement} shows the similarity map before and after interpolation.

\subsection{Objective Functions} \label{section:objective}
We utilize 4 different losses. 1) a location loss $L_\text{loc}$, 2) an angle loss $L_{\theta}$, 3) a contrastive loss $L_{\text{con}}$ of similarity among all anchors at the same level, 4) a reconstruction loss $L_2$ between the street-view PetalView feature and the best-matched satellite PetalView feature. Given ground truth $\{GT_x, GT_y, GT_\theta\}$, predicted result $\{P_x, P_y, P_\theta\}$, the similarity at each anchor $\{\phi_{N_\text{anchor}}\}$ and the cropped fine-grained matched PetalView features from satellite and street-view $F_\text{fine}^s$, $F_\text{fine}^g$. The objective function at each level is defined as:
\begin{equation}
\small
\label{eq:objectives}
\begin{split}
L = \lambda_\text{loc} L_\text{loc}  + \lambda_\theta L_\theta + &\lambda_\text{con} L_\text{con} + \lambda_\text{L2} L_2, \\
L_\text{loc} = \|\text{GT}_x-P_x, \text{GT}_y-P_y\|_{2}, & \: L_\text{con} = -\text{log}\frac{\exp(\phi^+/T)}{\sum_{i}^{N_\text{anchor}}\exp(\phi_i/T)} ,\\
L_2 =  \|F_\text{fine}^s - F_\text{fine}^g\|_2  ,\; L_\theta = |180^{\circ}& - ||\text{GT}_\theta - P_\theta | - 180^{\circ} ||/180^{\circ}, 
\end{split}
\end{equation}
where we set $T=0.05$, $\lambda_\text{loc} = 1$, $\lambda_\theta=1$, $\lambda_\text{con}=5$ and $\lambda_\text{L2}=0.2$ to weigh different loss functions for this work. The network is updated at the end of each mini-batch, instead of at every search level.

\section{Experiments}
\subsection{Datasets and Evaluation}
\noindent\textbf{Datasets.}
We test on two datasets, the selected KITTI-Satellite dataset~\cite{GeigerEtAl2013, ShiLi2022} and the VIGOR dataset~\cite{ZhuEtAl2021a, LentschEtAl2022}. \textbf{KITTI}~\cite{GeigerEtAl2013} contains street-view images with FOV of around $80^\circ$ and calibrated location/orientation from GPS/IMU. Since, the street-view images are upright, only the viewing direction in azimuth needs to be recovered. The corresponding satellite images are collected by \cite{ShiLi2022} with the GPS location of the street-view images and downloaded from Google Maps \cite{Gmaps} with a resolution of [1240, 1240] pixels. Different from the \cite{ShiLi2022}, we use the Earth-centric setting and a coverage of satellite images of [1024, 1024] pixels to extend the viewing range. The Kitti-Satellite dataset contains a train set, and two test sets - one in the same area as training and one in another area. The \textbf{VIGOR} dataset contains 360$^\circ$ street-view images and satellite images of 4 cities sampled with a 50\% overlap rate. As the satellite images are evenly sampled across the city, the camera locations of the street-view images are usually off the satellite image centers. Each street-view image has 1 positive and 3 semi-positive satellite samples. Following SliceMatch\cite{LentschEtAl2022}, We use the positive satellite sample and the updated labels. The street-view images are upright and north-aligned. We created misalignment by rotating the street-view images following \cite{HuEtAl2022a}. The satellite images are resized to [512, 512] and processed in Earth-centric setting. We follow the splits in \cite{ZhuEtAl2021a} for same area and cross area tests.

\noindent\textbf{Evaluation.}
For KITTI, following \cite{ShiLi2022}, we set location shift with a max limit of 20 meters in longitude and latitude direction. For orientation ground truth, we keep the 1) prior angle with $10^\circ$ absolute noise level (total $20^\circ$) and include the additional scenarios: 2) $20^\circ$ absolute noise level, 3) $40^\circ$ absolute noise level, 3) no prior angle/$180^\circ$ absolute noise level. For VIGOR dataset, we test on no prior angle setting and utilize the original off-center shift of the dataset within the center half of the satellite images. We evaluate the results with the recall within a threshold (r@x) and mean error $\epsilon_\text{Mean}$ on four aspects: longitude shift, latitude shift, orientation, and overall location shift. We report additional median error $\epsilon_\text{Med}$ to be aligned with the previous works on VIGOR dataset.

\subsection{Implementation Details}
We set the number of zones $N_z$ as 4 with the boundaries as [8/5, 20/9, 34/13, 48/17] meters from the anchors for KITTI/VIGOR. The angle width of the 4-level search is set as [10$^\circ$, 5$^\circ$, 2.5$^\circ$, 2.5$^\circ$] for KITTI and the 3-level search [10$^\circ$, 5$^\circ$, 2.5$^\circ$] for VIGOR. Our model PetalView is trained in the Earth-centric setting using Adam optimizer~\cite{KingmaB14} with an initial learning rate of $5e^{-5}$/$8e^{-4}$ for KITTI/VIGOR, a cosine learning rate scheduler with warm-up, cycle decay 0.8 and cycle length multiplier of 2. Early stopping of 20 epochs is applied on the train set. For KITTI, we re-implemented LM model from \cite{ShiLi2022} as it is the only previous work on this dataset with released code. LM model uses a car-centric setting, which aligns the satellite images direction to the noisy angle priors. This reduces the search difficulty when the noisy angle prior has a low discrepancy to the ground truth. Hence, we train the LM model in two modes: 1) directly in the Earth-centric setting with/without angle prior.\footnote{For no prior test, LM still requires an initial prior. The east direction is set as the prior.} 2) wrapping the input into the car-centric setting for prediction and unwrapping the results back to the Earth-centric setting for comparison. In the result section, the two modes are indicated as `ec' and `cc'. LM in \cite{ShiLi2022} was trained for 2 epochs, we train it with its original setting up to 5 epochs and take the best epoch. We apply random rotation on the satellite imagery as image augmentation for the models processed in Earth-centric settings to reduce the overfitting problem. As the code of SliceMatch~\cite{LentschEtAl2022} is not available yet, we reprint the VIGOR results reported in their paper.\footnote{SliceMatch is not compared on KITTI, as we are not able to verify their results yet.} LM does not work on VIGOR as the required camera parameters are not provided.

\section{Results and Discussion}
\begin{table*}[]
\small
\caption{Performance on VIGOR dataset with no angle prior.}
\vspace{-3mm}
\label{tab: vigor}
\begin{tabular}{|l|rrrr|rrrr|}
\hline
                    & \multicolumn{4}{c|}{Same Area}                                                                                                              & \multicolumn{4}{c|}{Cross Area}                                                                                                            \\ \cline{2-9} 
                    & \multicolumn{2}{c|}{Orientation ($^\circ$)}                                & \multicolumn{2}{c|}{Location (m)}                          & \multicolumn{2}{c|}{Orientation ($^\circ$)}                          & \multicolumn{2}{c|}{Location (m)}                          \\ \cline{2-9} 
                    & \multicolumn{1}{c|}{$\epsilon_\text{Mean} \downarrow$ }           & \multicolumn{1}{c|}{$\epsilon_\text{Med} \downarrow$ }        & \multicolumn{1}{c|}{$\epsilon_\text{Mean} \downarrow$ }          & \multicolumn{1}{c|}{$\epsilon_\text{Med} \downarrow$} & \multicolumn{1}{c|}{$\epsilon_\text{Mean} \downarrow$ }           & \multicolumn{1}{c|}{$\epsilon_\text{Med} \downarrow$}        & \multicolumn{1}{c|}{$\epsilon_\text{Mean} \downarrow$ }          & \multicolumn{1}{c|}{$\epsilon_\text{Med} \downarrow$} \\ \hline
MCC\cite{XiaEtAl2022}                 & \multicolumn{1}{r|}{56.86}          & \multicolumn{1}{r|}{16.02}         & \multicolumn{1}{r|}{9.87}          & 6.25                        & \multicolumn{1}{r|}{72.13}          & \multicolumn{1}{r|}{29.97}          & \multicolumn{1}{r|}{12.66}         & 9.55                        \\ \hline
SliceMatch\cite{LentschEtAl2022} VGG16    & \multicolumn{1}{r|}{28.43}          & \multicolumn{1}{r|}{5.15}          & \multicolumn{1}{r|}{8.41}          & 5.07                        & \multicolumn{1}{r|}{26.20}           & \multicolumn{1}{r|}{5.18}          & \multicolumn{1}{r|}{8.48}          & 5.64                        \\ \hline
SliceMatch\cite{LentschEtAl2022} ResNet50 & \multicolumn{1}{r|}{25.46}          & \multicolumn{1}{r|}{4.71}          & \multicolumn{1}{r|}{6.49}          & 3.13                        & \multicolumn{1}{r|}{25.97}          & \multicolumn{1}{r|}{4.51}          & \multicolumn{1}{r|}{7.22}          & 3.31                        \\ \hline
Ours                & \multicolumn{1}{r|}{\textbf{19.15}} & \multicolumn{1}{r|}{\textbf{1.78}} & \multicolumn{1}{r|}{\textbf{4.58}} & \textbf{1.71}               & \multicolumn{1}{r|}{\textbf{24.09}} & \multicolumn{1}{r|}{\textbf{2.03}} & \multicolumn{1}{r|}{\textbf{5.91}} & \textbf{1.97}               \\ \hline
\end{tabular}
\end{table*}

\begin{table*}[]
\small
\vspace{-3mm}
\caption{Performance on KITTI with no angle prior.}
\vspace{-3mm}
\label{tab:no prior}
\begin{tabular}{|p{1.6cm}|p{0.7cm}<{\centering}cp{0.4cm}<{\centering}ccp{0.5cm}<{\centering}|p{0.7cm}<{\centering}cp{0.4cm}<{\centering}p{0.7cm}<{\centering}cp{0.7cm}<{\centering}|}
\hline
\multicolumn{1}{|c|}{\multirow{2}{*}{\begin{tabular}[c]{@{}c@{}}Model \\ Name\end{tabular}}} & \multicolumn{3}{c|}{Lat}                                                                 & \multicolumn{3}{c|}{Lon}                                                                 & \multicolumn{3}{c|}{Orientation}                                                          & \multicolumn{3}{c|}{Loc}                                            \\  
\multicolumn{1}{|c|}{}                                                                       & \multicolumn{1}{c}{r@1m$\uparrow$}    & \multicolumn{1}{c}{r@5m$\uparrow$}    & \multicolumn{1}{c|}{$\epsilon_\text{Mean} \downarrow$}  & \multicolumn{1}{c}{r@1m$\uparrow$}    & \multicolumn{1}{c}{r@5m$\uparrow$}    & \multicolumn{1}{c|}{$\epsilon_\text{Mean} \downarrow$}  & \multicolumn{1}{c}{r@1$^\circ$$\uparrow$}    & \multicolumn{1}{c}{r@5$^\circ$$\uparrow$}    & \multicolumn{1}{c|}{$\epsilon_\text{Mean} \downarrow$}   & \multicolumn{1}{c}{r@1m}    & \multicolumn{1}{c}{r@5m}    & $\epsilon_\text{Mean} \downarrow$  \\ \hline
                                                                                             & \multicolumn{12}{c|}{Test 1}                               \\ \hline
LM\_ec\cite{ShiLi2022}                                                                                       & \multicolumn{1}{c}{7.61\%}  & \multicolumn{1}{c}{35.65\%} & \multicolumn{1}{c|}{9.06}  & \multicolumn{1}{c}{5.17\%}  & \multicolumn{1}{c}{25.97\%} & \multicolumn{1}{c|}{10.01} & \multicolumn{1}{c}{0.82\%}  & \multicolumn{1}{c}{4.24\%}  & \multicolumn{1}{c|}{84.30}  & \multicolumn{1}{c}{0.42\%}  & \multicolumn{1}{c}{9.01\%}  & 14.76 \\ \hline
Ours                                                                                         & \multicolumn{1}{c}{\textbf{79.35\%}} & \multicolumn{1}{c}{\textbf{94.94\%}} & \multicolumn{1}{c|}{\textbf{1.54}}  & \multicolumn{1}{c}{\textbf{81.50\%}} & \multicolumn{1}{c}{\textbf{94.99\%}} & \multicolumn{1}{c|}{\textbf{1.48}}  & \multicolumn{1}{c}{\textbf{21.10\%}} & \multicolumn{1}{c}{\textbf{78.21\%}} & \multicolumn{1}{c|}{\textbf{8.86}}   & \multicolumn{1}{c}{\textbf{68.88\%}} & \multicolumn{1}{c}{\textbf{92.84\%}} & \textbf{2.32}  \\ \hline

                                                                                             & \multicolumn{12}{c|}{Test 2}                      \\ \hline
LM\_ec\cite{ShiLi2022}                                                                                       & \multicolumn{1}{c}{8.15\%}  & \multicolumn{1}{c}{36.41\%} & \multicolumn{1}{c|}{\textbf{8.75}}  & \multicolumn{1}{c}{5.53\%}  & \multicolumn{1}{c}{25.76\%} & \multicolumn{1}{c|}{\textbf{9.98}}  & \multicolumn{1}{c}{0.53\%}  & \multicolumn{1}{c}{2.51\%}  & \multicolumn{1}{c|}{103.57} & \multicolumn{1}{c}{0.40\%}  & \multicolumn{1}{c}{9.86\%}  & \textbf{14.43} \\ \hline
Ours                                                                                         & \multicolumn{1}{c}{\textbf{13.72\%}} & \multicolumn{1}{c}{\textbf{37.39\%}} & \multicolumn{1}{c|}{14.85} & \multicolumn{1}{c}{\textbf{12.19\%}} & \multicolumn{1}{c}{\textbf{32.98\%}} & \multicolumn{1}{c|}{15.89} & \multicolumn{1}{c}{\textbf{10.21\%}} & \multicolumn{1}{c}{\textbf{37.72\%}} & \multicolumn{1}{c|}{\textbf{57.10}}  & \multicolumn{1}{c}{\textbf{5.82\%}}  & \multicolumn{1}{c}{\textbf{22.17\%}} & 23.79 \\ \hline
\end{tabular}
\end{table*}

\begin{table*}[]
\small
\vspace{-2mm}
\caption{Performance on KITTI in test 1 set for model trained with pre-defined angle prior noise level.}
\vspace{-3mm}
\label{tab:prior test1}
\begin{tabular}{|p{1.6cm}|ccc|ccc|ccc|ccc|}
\hline
        \multicolumn{1}{|c|}{\multirow{2}{*}{\begin{tabular}[c]{@{}c@{}}Model\\ Name\end{tabular}}}  & \multicolumn{3}{c|}{Lat}                                                                      & \multicolumn{3}{c|}{Lon}                                                                      & \multicolumn{3}{c|}{Orientation}                                                              & \multicolumn{3}{c|}{Loc}                                                                      \\  
           & \multicolumn{1}{c}{r@1m$\uparrow$}             & \multicolumn{1}{c}{r@5m$\uparrow$}             & $\epsilon_\text{Mean} \downarrow$          & \multicolumn{1}{c}{r@1m$\uparrow$}             & \multicolumn{1}{c}{r@5m$\uparrow$}             & $\epsilon_\text{Mean} \downarrow$          & \multicolumn{1}{c}{r@1$^\circ$$\uparrow$}             & \multicolumn{1}{c}{r@5$^\circ$$\uparrow$}             & $\epsilon_\text{Mean} \downarrow$         & \multicolumn{1}{c}{r@1m$\uparrow$}             & \multicolumn{1}{c}{r@5m$\uparrow$}             & $\epsilon_\text{Mean} \downarrow$          \\ \hline
LM\_40$^\circ$\_ec\cite{ShiLi2022} & \multicolumn{1}{c}{5.70\%}           & \multicolumn{1}{c}{28.60\%}          & 9.71          & \multicolumn{1}{c}{5.41\%}           & \multicolumn{1}{c}{28.23\%}          & 9.75          & \multicolumn{1}{c}{2.81\%}           & \multicolumn{1}{c}{12.62\%}          & 20.28         & \multicolumn{1}{c}{0.08\%}           & \multicolumn{1}{c}{6.76\%}           & 15.01         \\ 
LM\_40$^\circ$\_cc\cite{ShiLi2022} & \multicolumn{1}{c}{11.45\%}          & \multicolumn{1}{c}{43.04\%}          & 8.06          & \multicolumn{1}{c}{12.72\%}          & \multicolumn{1}{c}{47.97\%}          & 7.62          & \multicolumn{1}{c}{4.72\%}           & \multicolumn{1}{c}{19.56\%}          & 17.63         & \multicolumn{1}{c}{1.64\%}           & \multicolumn{1}{c}{19.98\%}          & 12.44         \\ 
Ours\_40$^\circ$   & \multicolumn{1}{c}{\textbf{81.34\%}} & \multicolumn{1}{c}{\textbf{95.34\%}} & \textbf{1.45} & \multicolumn{1}{c}{\textbf{82.93\%}} & \multicolumn{1}{c}{\textbf{95.31\%}} & \textbf{1.49} & \multicolumn{1}{c}{\textbf{21.57\%}} & \multicolumn{1}{c}{\textbf{81.26\%}} & \textbf{4.33} & \multicolumn{1}{c}{\textbf{71.51\%}} & \multicolumn{1}{c}{\textbf{93.69\%}} & \textbf{2.26} \\ \hline
LM\_20$^\circ$\_ec\cite{ShiLi2022} & \multicolumn{1}{c}{5.38\%}           & \multicolumn{1}{c}{25.34\%}          & 10.44         & \multicolumn{1}{c}{6.49\%}           & \multicolumn{1}{c}{27.48\%}          & 10.21         & \multicolumn{1}{c}{5.59\%}           & \multicolumn{1}{c}{25.95\%}          & 10.16         & \multicolumn{1}{c}{0.19\%}           & \multicolumn{1}{c}{6.33\%}           & 15.87         \\ 
LM\_20$^\circ$\_cc\cite{ShiLi2022} & \multicolumn{1}{c}{11.85\%}          & \multicolumn{1}{c}{46.86\%}          & 7.80          & \multicolumn{1}{c}{13.78\%}          & \multicolumn{1}{c}{49.70\%}          & 7.57          & \multicolumn{1}{c}{\textbf{25.66\%}} & \multicolumn{1}{c}{70.82\%}          & 4.91          & \multicolumn{1}{c}{1.88\%}           & \multicolumn{1}{c}{22.05\%}          & 12.22         \\ 
Ours\_20$^\circ$   & \multicolumn{1}{c}{\textbf{82.48\%}} & \multicolumn{1}{c}{\textbf{95.89\%}} & \textbf{1.37} & \multicolumn{1}{c}{\textbf{84.18\%}} & \multicolumn{1}{c}{\textbf{95.57\%}} & \textbf{1.38} & \multicolumn{1}{c}{22.08\%}          & \multicolumn{1}{c}{\textbf{81.74\%}} & \textbf{3.94} & \multicolumn{1}{c}{\textbf{73.10\%}} & \multicolumn{1}{c}{\textbf{94.17\%}} & \textbf{2.10} \\ \hline
LM\_10$^\circ$\_ec\cite{ShiLi2022} & \multicolumn{1}{c}{6.71\%}           & \multicolumn{1}{c}{30.21\%}          & 10.05         & \multicolumn{1}{c}{5.99\%}           & \multicolumn{1}{c}{29.50\%}          & 10.49         & \multicolumn{1}{c}{9.33\%}           & \multicolumn{1}{c}{49.70\%}          & 5.11          & \multicolumn{1}{c}{0.37\%}           & \multicolumn{1}{c}{8.45\%}           & 15.94         \\ 
LM\_10$^\circ$\_cc\cite{ShiLi2022} & \multicolumn{1}{c}{12.32\%}          & \multicolumn{1}{c}{47.10\%}          & 7.65          & \multicolumn{1}{c}{14.10\%}          & \multicolumn{1}{c}{49.72\%}          & 7.42          & \multicolumn{1}{c}{\textbf{35.83\%}} & \multicolumn{1}{c}{81.69\%}          & \textbf{2.88} & \multicolumn{1}{c}{2.15\%}           & \multicolumn{1}{c}{23.40\%}          & 11.99         

\\ 
Ours\_10$^\circ$   & \multicolumn{1}{c}{\textbf{79.62\%}} & \multicolumn{1}{c}{\textbf{94.78\%}} & \textbf{1.59} & \multicolumn{1}{c}{\textbf{82.61\%}} & \multicolumn{1}{c}{\textbf{94.51\%}} & \textbf{1.60} & \multicolumn{1}{c}{26.50\%}          & \multicolumn{1}{c}{\textbf{88.60\%}} & 4.18          & \multicolumn{1}{c}{\textbf{70.05\%}} & \multicolumn{1}{c}{\textbf{92.92\%}} & \textbf{2.44} \\ \hline
\end{tabular}
\end{table*}

\begin{table*}[h]
\small

\vspace{-3mm}
\caption{Ablation study on multi-scale vs flat search, viewing range and refinement strategies.}
\label{tab:ablation1}
\vspace{-3mm}
\begin{tabular}{|c|c|c|cccccc|cccccc|}
\hline
\multirow{3}{*}{Search} & \multirow{3}{*}{Size} & \multirow{3}{*}{Refine} & \multicolumn{6}{c|}{Test 1}                                                    & \multicolumn{6}{c|}{Test 2}                                                    \\ \cline{4-15} 
  &   &                         & \multicolumn{3}{c|}{Orientation}                   & \multicolumn{3}{c|}{Location} & \multicolumn{3}{c|}{Orientation}                   & \multicolumn{3}{c|}{Location} \\               &     &                         & r@1$^\circ$$\uparrow$    & r@5$^\circ$$\uparrow$    & \multicolumn{1}{c|}{$\epsilon_\text{Mean} \downarrow$}  & r@1m$\uparrow$      & r@5m$\uparrow$      & $\epsilon_\text{Mean} \downarrow$  & r@1$^\circ$$\uparrow$    & r@5$^\circ$$\uparrow$    & \multicolumn{1}{c|}{$\epsilon_\text{Mean} \downarrow$}  & r@1m$\uparrow$     & r@5m$\uparrow$      & $\epsilon_\text{Mean} \downarrow$   \\ \hline
 \textbf{Multi-scale} & \textbf{1024}                  & \textbf{Always}                  & 21.10\% & 78.21\% & \multicolumn{1}{c|}{\textbf{8.86}}  & \textbf{68.88\%}   & \textbf{92.84\%}   & \textbf{2.32}  & \textbf{10.21\%} & \textbf{37.72\%} & \multicolumn{1}{c|}{57.10} & 5.82\%   & 22.17\%   & 23.79  \\ \cline{1-3}

 Flat & 1024                  & None                & 8.06\% & 34.88\% & \multicolumn{1}{c|}{29.29}  & 2.15\%   & 29.34\%   & 11.05  & 4.97\% & 25.36\% & \multicolumn{1}{c|}{\textbf{49.29}} & 0.85\%   & 12.21\%   & 22.47  \\\cline{1-3}

 Multi-scale & 512                   & Always                   & 17.10\% & 71.51\% & \multicolumn{1}{c|}{9.73} & 44.24\%   & 91.86\%   & 2.57 & 8.47\%  & 35.96\% & \multicolumn{1}{c|}{49.89} & \textbf{6.27\%}   & \textbf{36.32\%}   & \textbf{13.54}  \\ \cline{1-3}
 
 Multi-scale & 1024                  & Last                    & \textbf{22.34\%} & 79.51\% & \multicolumn{1}{c|}{11.51}  & 63.50\%   & 91.41\%   & 2.96  & 9.18\% & 32.41\% & \multicolumn{1}{c|}{67.24} & 5.13\%   & 22.33\%   & 25.19  \\   \cline{1-3}
 Multi-scale & 1024                  & None                    & 21.39\% & \textbf{80.68\%} & \multicolumn{1}{c|}{10.76}  & 64.59\%   & 90.91\%  & 2.92  & 9.45\%  & 33.52\% & \multicolumn{1}{c|}{64.78} & 5.17\%   & 21.97\%   & 24.68  \\ 
\hline
\end{tabular}
\end{table*}

\subsection{Extraction without a Prior Orientation}
The most generic use case for local area search is to have a location prior without an angle prior. \autoref{tab: vigor} shows the results on VIGOR tests. We reprint the results of MCC~\cite{XiaEtAl2022} and SliceMatch~\cite{LentschEtAl2022} reported in \cite{LentschEtAl2022}. For all reported metrics in the previous paper, we obtain the best performance, achieving mean error on orientation and location 19.15$^\circ$/24.09$^\circ$ and 4.58/5.91 meters for the two tests. MCC focuses on localization, their orientation extraction is done by testing multiple rotated copies of the street-view image. The granularity of orientation estimation is depending on the number of copies tested, which is not ideal for efficient processing. Compared to SliceMatch, which only use orientation cue for feature gathering, our PetalView features leverage both orientation and relative distance to restrict the available information and structure a more delicate feature representation. With the multi-scale search, we are able to process larger feature maps with adjustable granularity at different levels to ensure fine granularity with a reduced computational cost compared to flat search. Hence, our model shows significantly better performance than SliceMatch. We also report the r@1m/1$^\circ$ and other metrics in \autoref{tab:Add Vigor} in Appendix~\ref{apex: add vigor} as a reference for future studies.

\autoref{tab:no prior} shows the results on KITTI dataset. LM~\cite{ShiLi2022} is designed to refine the location and orientation when the initial camera pose is not far from the ground truth. Without the angle prior, \ie noise level to 180$^\circ$, LM does not perform well and barely improved on the initial error (test 1 init: 4.88\%, 4.77\%, 0.56\% and 0.16\% for r@1m/1$^\circ$ on lat, lon, orientation and location). Our PetalView model significantly improves performance on all metrics in the same area test~1 compared to LM. The r@1m for lat, lon and overall location reach 79.35\%, 81.50\%, 68.88\% respectively, and r@1$^\circ$ for orientation reaches 21.10\%. It shows our model does not depend on an angle prior to search through the entire 3D search space. Most of the metrics, besides the location-related mean errors, in the cross-area test~2 also improved, but this improvement is not as significant as in test 1. We observe there is a performance gap between the two tests, which is not obvious in VIGOR. 

We hypothesize that the KITTI dataset more easily lead to a performance gap than VIGOR because 1) KITTI street-view images have a lower FOV of 80$^\circ$, which provide less surrounding information; 2) KITTI are densely sampled along few tracks, while VIGOR samples the whole cities and undergoes a balancing procedure to ensure the sparsity of the samples, which let the two test sets have more similar natures than the two sets in KITTI. Compared to existing methods, our model achieves significantly better performance in VIGOR and KITTI test~1. On KITTI test 2, we observe performance gaps to test 1 and our model leans to the training area when the FOV is limited. This is one of the aspects to be improved in the future study.

\subsection{Extraction with a Prior Orientation}\label{section: with prior}
An angle prior can be leveraged to guide the search. \autoref{tab:prior test1} and \autoref{tab:prior test2} (Appendix~\ref{appex: kitti 2 prior}) show the results on KITTI tests. We observe that: 1) After adding the learnable prior parameters in the model, all metrics have similar or better performance compared to no angle prior setting (\autoref{tab:no prior}). 2) Compared to LM, our model has significantly improved location estimation for all noise levels. For orientation estimation, as LM car-centric wrapping process rotates the satellite imagery to the heading direction, it is expected to observe a higher performance at lower noise levels. The r@1$^\circ$ at 10$^\circ$ and 20$^\circ$ noise levels obtain better performance than ours, but the gaps decrease with the increased noise levels. At 40$^\circ$ noise level, our model still holds a decent r@1$^\circ$ 21.57\%, while LM gives 4.72\%, which barely improves over the initial value of 2.41\% for r@1$^\circ$. Our multi-scale search strategy significantly increases the granularity of the location estimation, even when the orientation estimation has a similar level of performance. 3) Comparing the results of our model across different noise levels, the performance of location estimation obtains better performance than without angle prior, and the orientation estimation accuracy improves when the noise level is decreased for both tests. It shows our prior angle mixer can effectively adapt the prior angle to the original estimation curve. And 4) with this dynamically learnable prior mixer, the orientation gaps between the two tests also reduce, which means the knowledge of angle prior is transferable and helps to reduce overfitting issues.

\subsection{Ablation Study on KITTI}
\subsubsection{Multi-scale search}
To study the impact brought by multi-scale search, we train a comparison model requiring slightly higher computational costs but with a flat regular search grid. Our result used 4-level search with [16, 16, 16, 9] anchors and viewing angles of [10$^\circ$, 5$^\circ$, 2.5$^\circ$, 2.5$^\circ$] at each level. The comparison model uses a 64-anchor regular grid in inference and 16 randomly selected anchors with 4 times the learning rate in training, the viewing angle is fixed at 2.5$^\circ$. As the randomly selected anchors are not optimized for location refinement, we compare the flat grid results (row 2) with no location refinement (row 5 in \autoref{tab:ablation1}). Our multi-scale search strategy indeed efficiently improves the searching granularity and provides a significant performance improvement. Moreover, if we compare our flat search result with the LM model (\autoref{tab:no prior}), our model still has better performance, which shows even without multi-scale search our PetalView network brought significant improvement.
\vspace{-2mm}
\subsubsection{Viewing range}
Viewing range refers to how far the PetalView feature observes. The main results on KITTI use zone boundaries of [8, 20, 34, 48] meters. An alternative setting is at [6, 12, 18, 24] meters for satellite images with 512$\times$ 512 pixels
with a 3-level search. Row 1 and 3 in \autoref{tab:ablation1} show the results. When the viewing range is shorter, the orientation r@1$^\circ$ for both test sets obtains worse performance. Given that street-view images in an urban setting can reference objects at a distance of 50 meters or further, our interpretation is that far away landmarks can be good features to match the two views. For location estimation, the model using a short view has a lower performance in test~1 but a higher performance in test~2, which shows that with direct supervision in the same area, having a larger viewing range still benefits location estimation although the search area increases when the input size increases. However, the increased viewing range does not overweigh the enlarged search area when there is no training sample in the test area.
\subsubsection{Location refinement}
Interpolation on the similarity scores of anchors helps to find the sub-anchor location. Row~1,4 and 5 of \autoref{tab:ablation1} show the results trained with different refinement strategies. Always: every level; last: last level only; none: no refinement. The results show that the refinement strategy not only affects the performance of the location extraction but also the orientation extraction, as the precision of the location error influences the balance of the losses. Refining only the last level may bring inconsistency across all levels. `Always' gives the best overall results.

\section{Conclusions}

In this work, we focus on finding fine-grained orientation and location efficiently in local-scale search. To improve the granularity of the estimation and to reduce the number of queries on the satellite image, we propose PetalView feature extractors with multi-scale search strategy. To cooperate with the angle prior information, a learnable prior curve is introduced to adapt to the influence of the prior. Our proposed model obtained the best performance on VIGOR and successfully improves the location estimation on KITTI to about 69\%/6\% and orientation estimation to about 21\%/10\% for same/cross-area tests when no angle prior is available. When an angle prior is given, our method has stable performance that gives r@1m/r@1$^\circ$ above 70\%/21\% for test 1, above 6\%/13\% for test 2. Additionally, our model does not require information exchange between the two views till the final matching at each level, which makes it possible to extend usage to large-scale search. For future studies, we would like to reduce the gap between the same area test and cross area test to improve the transferable of the method.

\begin{acks}
This work was funded by the Grab-NUS AI Lab, a joint collaboration between GrabTaxi Holdings Pte. Ltd. and National University of Singapore, and the Industrial Postgraduate Program (Grant: S18-1198-IPP-II) funded by the Economic Development Board of Singapore. Y. Liang is supported by Guangzhou Municipal Science and Technology Project 2023A03J0011.
\end{acks}

\clearpage
\bibliographystyle{ACM-Reference-Format}
\balance
\bibliography{main}
\clearpage

\appendix
\section{Appendix}
\subsection{Algorithm for Dynamic Multi-scale Grid Generation} \label{appex: multi-scale}

 To reduce the number of queries required for obtaining pixel-level precision search in the feature map space, we propose multi-scale search in Section~\ref{section:multi-scale} that restructures classic flat search into multi-level hierarchical search with increased precision granularity at each level.
Algorithm~\ref{alg:multi-scale} shows how to generate next level anchors in the multi-scale search strategies.

\begin{algorithm}[ht!]
\caption{Dynamic multi-scale search}
\label{alg:multi-scale}
\small
\KwIn{Search area $L_s [H, W]$, grid side sample number $N_s$, grid side sample number at last level $N_s'$, search level $l \in [0, N_l)$, selection previous level $S_{l-1}$, grid locations $X$.}
\KwOut{Grid locations $X$.}
\uIf{l=0} {
      $X_c \gets \text{image center}$ \hfill // $X_c$ is the search center
    }
    \Else{
     $X_c \gets X[l-1] \hfill $ // $X_c$ as the previous session selection
    }

\uIf{\text{l not last level}}{
    $L_p \gets L_s/(N_s)^l$ ,\: $N_\text{anchor}=(N_s)^2$ \hfill // $L_p$ is patch size \\
    
    }
    \Else{
    $L_p \gets L_s/(N_s)^{(l-1)}$ ,\: $N_\text{anchor}=(N_s')^2$
    }

$X_l \gets \text{create\_patch\_centers}(X_c, L_p, N_\text{anchor})$ \hfill \\
$X.append(X_l)$

\Return{$X$}
\end{algorithm}

\subsection{Learnable Prior Curve} \label{section: appendix prior}

\begin{figure}[!b]
  \includegraphics[width=0.49\textwidth,trim = 0cm 0cm 0cm 0cm, clip]{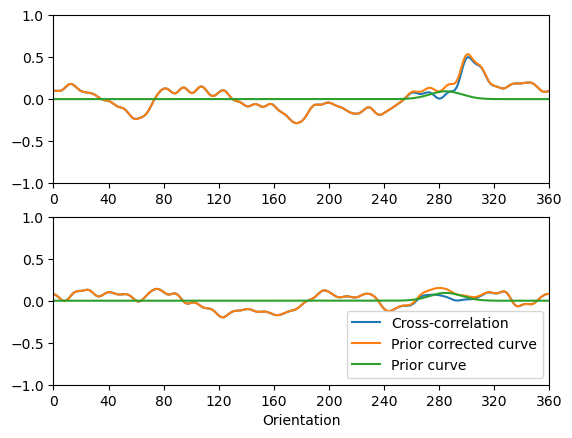}
  \caption{Similarity curve, prior angle information adds to the cross-correlation curve. Top: Curves from an anchor that is close to the correct location. Bottom: Curves from an anchor that is far away from the correct location.}
  \label{fig:prior_curve}
\end{figure}
When prior orientation information is given, the MatchMaker has additional learnable parameters to control the influence of the prior. \autoref{fig:prior_curve} shows the influence of prior on two cases.
The prior curve has a stronger influence on the final orientation extraction when the overall similarity curve is low and only gives minor adjustments to the curve when the overall similarity curve has a high score. Through learning the similarity curves of all positive and negative samples of the dataset, the model learns reasonable parameters of the prior curve that balance the guidance of the prior.

\subsection{Refinement of Location} \label{section: refinement}
\begin{figure}[!h]
\vspace{-3mm}
  \includegraphics[width=0.49\textwidth,trim = 4.5cm 5.2cm 4.5cm 0.5cm, clip]{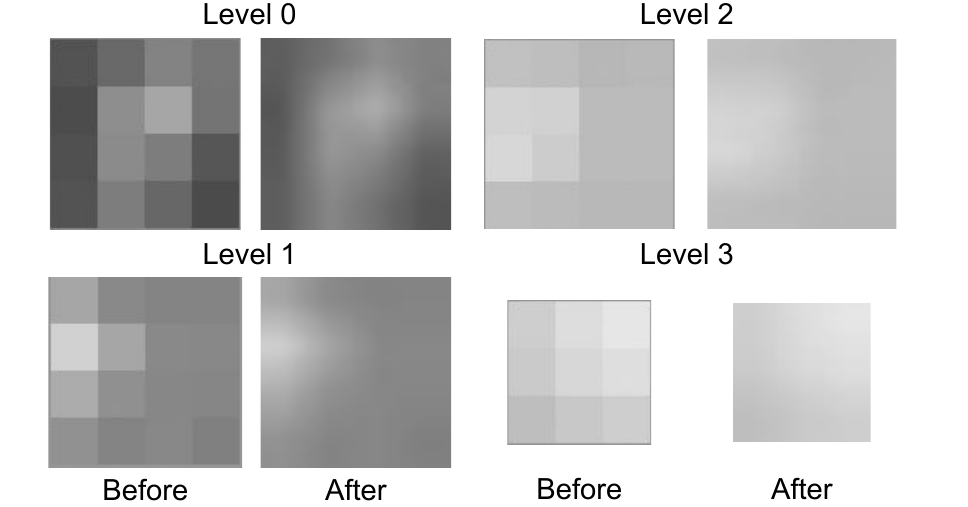}
  \caption{Similarity scores before and after location refinement at 4 search levels.}
  \label{fig:refine_location}
\end{figure}

\autoref{fig:refine_location} demonstrates the effect of location refinements to reach sub-anchor level search precision. For a 4-level search with anchor number [16, 16, 16, 9] on KITTI. The corresponding anchor distances are [$25.07, 6.27, 1.57, 0.78$] meters, which indicates the upper limit of the retrieval precision at each level without interpolation. By upsampling the similarity score 8 times, the smoothed similarity map has a maximum resolving power of [$3.13, 0.38, 0.20, 0.10$] meters at each level. The final output has a precision of 0.1 meter as the maximum resolving power.
\vspace{-3mm}

\begin{table*}[h]
\small
\vspace{-3mm}
\caption{Ablation study on petal information processing strategies.}

\label{tab:ablation2}
\vspace{-3mm}
\begin{tabular}{|c|c|cccccc|cccccc|}
\hline
\multirow{3}{*}{Satellite} & \multirow{3}{*}{Street} & \multicolumn{6}{c|}{Test 1}                                                    & \multicolumn{6}{c|}{Test 2}                                                    \\ \cline{3-14}                        &                         & \multicolumn{3}{c|}{Orientation}                   & \multicolumn{3}{c|}{Location} & \multicolumn{3}{c|}{Orientation}                   & \multicolumn{3}{c|}{Location} \\
&                         & r@1$^\circ$$\uparrow$    & r@5$^\circ$$\uparrow$    & \multicolumn{1}{c|}{$\epsilon_\text{Mean} \downarrow$}  & r@1m$\uparrow$      & r@5m$\uparrow$      & $\epsilon_\text{Mean} \downarrow$  & r@1$^\circ$$\uparrow$    & r@5$^\circ$$\uparrow$    & \multicolumn{1}{c|}{$\epsilon_\text{Mean} \downarrow$}  & r@1m$\uparrow$     & r@5m$\uparrow$      & $\epsilon_\text{Mean} \downarrow$   \\ \hline

\textbf{RAP}                        & \textbf{HAP}           & 21.10\% & 78.21\% & \multicolumn{1}{c|}{\textbf{8.86}}  & \textbf{68.88\%}   & \textbf{92.84\%}   & \textbf{2.32}  & 10.21\% & \textbf{37.72\%} & \multicolumn{1}{c|}{\textbf{57.10}} & \textbf{5.82\%}   & \textbf{22.17\%}   & 23.79  \\ \cline{1-2}

Pooling                    & Pooling     & 21.28\% & 77.05\% & \multicolumn{1}{c|}{16.04} & 67.00\%   & 89.88\%   & 2.52  & 7.52\%  & 29.66\% & \multicolumn{1}{c|}{59.48} & 4.27\%   & 16.10\%   & 25.55  \\ \cline{1-2}
AP                         & Pooling           & 
17.44\% & 67.27\% & \multicolumn{1}{c|}{27.79} & 52.43\%   & 88.60\%   & 3.24  & 7.62\% & 33.20\% & \multicolumn{1}{c|}{67.25} & 4.14\%   & 18.93\%   & \textbf{23.46}  \\ \cline{1-2}
RAP                        & DAP       & 19.72\% & 77.50\% & \multicolumn{1}{c|}{9.37} & 67.77\%   & 92.31\%   & 2.44  & \textbf{10.89\%} & 34.87\% & \multicolumn{1}{c|}{62.27} & 5.20\%   & 21.51\%   & 24.68  \\ \cline{1-2}
RAP                        & HCAP        & 19.19\% & 76.57\% & \multicolumn{1}{c|}{13.12} & 66.79\%   & 91.84\%   & 2.64  & 6.06\%  & 25.80\% & \multicolumn{1}{c|}{68.52} & 3.54\%   & 17.05\%   & 26.25  \\ \cline{1-2}
RAP                        & DCAP          & \textbf{23.48\%} & \textbf{78.80\%} & \multicolumn{1}{c|}{14.09} & 62.95\%   & 89.19\%   & 3.56  & 9.35\%  & 32.48\% & \multicolumn{1}{c|}{64.58} & 4.56\%   & 20.43\%   & 24.18  \\  
\hline
\end{tabular}
\vspace{-3mm}
\end{table*}

\begin{table*}[]
\small

\caption{Additional metrics on VIGOR dataset.}
\vspace{-3mm}
\label{tab:Add Vigor}
\begin{tabular}{|l|cccc|cccc|cc|cc|}
\hline
       & \multicolumn{4}{c|}{Lat (m)}                                                                                     & \multicolumn{4}{c|}{Lon (m)}                                                                                     & \multicolumn{2}{c|}{Orientation ($^\circ$)}                      & \multicolumn{2}{c|}{Location (m)}                        \\ \cline{2-13} 
       & \multicolumn{1}{l}{r@1m$\uparrow$} & \multicolumn{1}{l}{r@5m$\uparrow$} & \multicolumn{1}{l}{$\epsilon_\text{Mean} \downarrow$} & \multicolumn{1}{l|}{$\epsilon_\text{Med} \downarrow$} & \multicolumn{1}{l}{r@1m$\uparrow$} & \multicolumn{1}{l}{r@5m$\uparrow$} & \multicolumn{1}{l}{$\epsilon_\text{Mean} \downarrow$} & \multicolumn{1}{l|}{$\epsilon_\text{Med} \downarrow$} & \multicolumn{1}{l}{r@1$^\circ$$\uparrow$} & \multicolumn{1}{l|}{r@5$^\circ$$\uparrow$} & \multicolumn{1}{l}{r@1m$\uparrow$} & \multicolumn{1}{l|}{r@5m$\uparrow$} \\ \hline
Same area & 49.32\%                  & 85.26\%                  & 2.89                     & 1.01                        & 50.61\%                  & 85.56\%                  & 2.82                     & 0.97                        & 33.02\%                  & 76.14\%                   & 29.08\%                  & 76.90\%                   \\ \hline
Cross area & 45.73\%                  & 81.42\%                  & 3.58                     & 1.15                        & 45.95\%                  & 80.72\%                  & 3.67                     & 1.14                        & 30.10\%                  & 71.85\%                   & 24.19\%                  & 70.26\%                   \\ \hline
\end{tabular}
\end{table*}

\begin{table*}[]
\small

\vspace{-3mm}
\caption{Performance in KITTI test 2 set for model trained with pre-defined angle prior noise level.}
\vspace{-3mm}
\label{tab:prior test2}
\begin{tabular}{|p{1.6cm}|ccc|ccc|ccc|ccc|}
\hline
\multicolumn{1}{|c|}{\multirow{2}{*}{\begin{tabular}[c]{@{}c@{}}Model\\ Name\end{tabular}}} & \multicolumn{3}{c|}{Lat}                                                                      & \multicolumn{3}{c|}{Lon}                                                                      & \multicolumn{3}{c|}{Orientation}                                                               & \multicolumn{3}{c|}{Loc}                                                                      \\ 
\multicolumn{1}{|c|}{}                                                                      & \multicolumn{1}{c}{r@1m$\uparrow$}             & \multicolumn{1}{c}{r@5m$\uparrow$}             & $\epsilon_\text{Mean} \downarrow$          & \multicolumn{1}{c}{r@1m$\uparrow$}             & \multicolumn{1}{c}{r@5m$\uparrow$}             & $\epsilon_\text{Mean} \downarrow$         & \multicolumn{1}{c}{r@1$^\circ$$\uparrow$}             & \multicolumn{1}{c}{r@5$^\circ$$\uparrow$}             & $\epsilon_\text{Mean} \downarrow$          & \multicolumn{1}{c}{r@1m$\uparrow$}            & \multicolumn{1}{c}{r@5m$\uparrow$}             & $\epsilon_\text{Mean} \downarrow$          \\ \hline
LM\_40$^\circ$\_ec\cite{ShiLi2022}                                                                                  & \multicolumn{1}{c}{5.97\%}           & \multicolumn{1}{c}{28.31\%}          & 9.80          & \multicolumn{1}{c}{5.58\%}           & \multicolumn{1}{c}{27.00\%}          & 9.93          & \multicolumn{1}{c}{2.33\%}           & \multicolumn{1}{c}{12.13\%}          & 20.21          & \multicolumn{1}{c}{0.40\%}          & \multicolumn{1}{c}{6.46\%}           & 15.17          \\ 
LM\_40$^\circ$\_cc\cite{ShiLi2022}                                                                                  & \multicolumn{1}{c}{10.75\%}          & \multicolumn{1}{c}{42.43\%}          & \textbf{8.23} & \multicolumn{1}{c}{8.01\%}           & \multicolumn{1}{c}{37.21\%}          & \textbf{8.55} & \multicolumn{1}{c}{4.44\%}           & \multicolumn{1}{c}{20.27\%}          & \textbf{17.98} & \multicolumn{1}{c}{1.68\%}          & \multicolumn{1}{c}{18.83\%}          & \textbf{12.97} \\ 
Ours\_40$^\circ$                                                                                    & \multicolumn{1}{c}{\textbf{16.06\%}} & \multicolumn{1}{c}{\textbf{44.54\%}} & 12.83         & \multicolumn{1}{c}{\textbf{15.59\%}} & \multicolumn{1}{c}{\textbf{40.41\%}} & 13.90         & \multicolumn{1}{c}{\textbf{13.72\%}} & \multicolumn{1}{c}{\textbf{52.48\%}} & 22.13          & \multicolumn{1}{c}{\textbf{6.40\%}} & \multicolumn{1}{c}{\textbf{29.17\%}} & 20.76          \\ \hline
LM\_20$^\circ$\_ec\cite{ShiLi2022}                                                                                  & \multicolumn{1}{c}{4.72\%}           & \multicolumn{1}{c}{26.33\%}          & 10.46         & \multicolumn{1}{c}{5.44\%}           & \multicolumn{1}{c}{26.61\%}          & 10.19         & \multicolumn{1}{c}{5.13\%}           & \multicolumn{1}{c}{25.23\%}          & 10.21          & \multicolumn{1}{c}{0.20\%}          & \multicolumn{1}{c}{6.01\%}           & 15.88          \\ 
LM\_20$^\circ$\_cc\cite{ShiLi2022}                                                                                  & \multicolumn{1}{c}{10.36\%}          & \multicolumn{1}{c}{42.89\%}          & \textbf{8.40} & \multicolumn{1}{c}{8.35\%}           & \multicolumn{1}{c}{37.01\%}          & \textbf{8.86} & \multicolumn{1}{c}{\textbf{21.23\%}} & \multicolumn{1}{c}{\textbf{64.80\%}} & \textbf{5.81}  & \multicolumn{1}{c}{1.15\%}          & \multicolumn{1}{c}{18.81\%}          & \textbf{13.38} \\ 
Ours\_20$^\circ$                                                                                    & \multicolumn{1}{c}{\textbf{16.89\%}} & \multicolumn{1}{c}{\textbf{45.19\%}} & 12.12         & \multicolumn{1}{c}{\textbf{16.48\%}} & \multicolumn{1}{c}{\textbf{41.71\%}} & 13.48         & \multicolumn{1}{c}{15.66\%}          & \multicolumn{1}{c}{61.40\%}          & 12.92          & \multicolumn{1}{c}{\textbf{7.25\%}} & \multicolumn{1}{c}{\textbf{30.60\%}} & 19.92          \\ \hline
LM\_10$^\circ$\_ec\cite{ShiLi2022}                                                                                  & \multicolumn{1}{c}{6.52\%}           & \multicolumn{1}{c}{30.56\%}          & 9.80          & \multicolumn{1}{c}{5.56\%}           & \multicolumn{1}{c}{26.81\%}          & 10.29         & \multicolumn{1}{c}{10.97\%}          & \multicolumn{1}{c}{51.30\%}          & 5.01           & \multicolumn{1}{c}{0.27\%}          & \multicolumn{1}{c}{7.54\%}           & 15.54          \\ 
LM\_10$^\circ$\_cc\cite{ShiLi2022}                                                                                  & \multicolumn{1}{c}{11.16\%}          & \multicolumn{1}{c}{\textbf{42.96\%}}          & \textbf{8.20} & \multicolumn{1}{c}{9.12\%}           & \multicolumn{1}{c}{\textbf{39.09\%}}          & \textbf{8.55} & \multicolumn{1}{c}{\textbf{28.40\%}} & \multicolumn{1}{c}{\textbf{78.51\%}} & \textbf{3.26}  & \multicolumn{1}{c}{1.78\%}          & \multicolumn{1}{c}{20.64\%}          & \textbf{12.96} 

\\ 
Ours\_10$^\circ$                                                                                    & \multicolumn{1}{c}{\textbf{15.77\%}} & \multicolumn{1}{c}{42.35\%} & 12.81         & \multicolumn{1}{c}{\textbf{14.21\%}} & \multicolumn{1}{c}{37.30\%} & 14.98         & \multicolumn{1}{c}{18.84\%}          & \multicolumn{1}{c}{57.15\%}          & 33.33          & \multicolumn{1}{c}{\textbf{6.18\%}} & \multicolumn{1}{c}{\textbf{27.58\%}} & 21.55          \\ \hline
\end{tabular}
\vspace{-3mm}
\end{table*}

\subsection{Visualization of Petal Sampling} \label{section:level0 samples}
\begin{figure}[]
  \includegraphics[width=0.45\textwidth,trim = 6.5cm 2cm 10.5cm 1cm, clip]{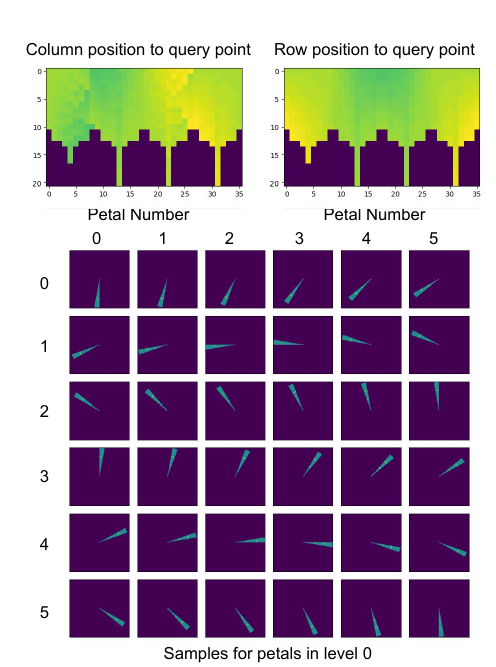}
  \caption{Lookup table and petal visualization for level 0.}
  \label{fig:level0_sample}
\end{figure}

\autoref{fig:level0_sample} top shows the column and row position of the selected pixels for each petal at level 0 zone 0 relative to the center of the anchor. The x-axis represents the 36 petals in level 0, when the width of viewing angle $\theta_a$ is set to 10$^\circ$; the y-axis represents the index of pixels in each petal. As each petal has a different alignment with the regular grid of the image, some petals have more pixels than others. To have a regular lookup table, we pad the length of the pixel number to the maximum pixel number at each zone at each level. The padded pixels are shown with dark blue color with default values. The bottom \autoref{fig:level0_sample} shows the visualization of the selected pixels for each petal in level 0.

\subsection{Ablation on Feature-processing Strategies}
After sampling pixels for each petal, different processes can be applied to obtain the PetalView features. \autoref{tab:ablation2} shows 6 different processing strategies. RAP: RangeAngle Processor; HAP: HeightAngle processor; Pooling: average pooling without zones, all pixels in the same viewing angle are pooled into one feature; AP: Angle Processor without zones; DAP: DepthAngle Processor, a depth estimator is added to find the distance of the pixels of the street-view images. The reversed depth discrepancy is used in the embedding instead of the height position in the column; HCAP and DCAP: Height-based Column Processor + Angle Processor and Depth-based Column Processor + Angle Processor. A column processor is a transform-based network like the HeightAngle processor but outputs zone-based features for each column as an intermediate result. When searching at each level, a transformer-based angle processor takes the relevant column feature to produce the final PetalView feature. The column+angle processor disentangles the zone-based process and angle-based process, which leads to all levels sharing the same zone-based intermediate results. 

We observe: 1) When using only the average pooling without zoning, the performance in test~1 still holds, but it has a worse performance gap in test~2. It means the model is too simple and only learns features that are less transferable. 2) After switching to AP without zoning, the gaps between the two tests are shortened. However, losing the zone information crucially affects the performance compared to the main results in row 1. 3) When changing the HAP to DAP, the test~2 orientation prediction is better, but the location prediction is worse as well as the orientation for test~1. We believe without depth ground truth, the learned distances are not accurate and do not help in improving the performance. And 4) if we separate the zone-based process and the angle-based process, the results are shown in row 5-6. The model with column processor and depth estimation shows a clear improvement in orientation estimation in test~1. However, as the image is processed line by line and there is a clear boundary between each fine-grained angle, this leads to another problem that the information between the columns may be discontinuous. It lowers the overall quality of the features which affects transferability and fails to give the best-fit anchor a distinct high score.
\subsection{Additional Test Results on VIGOR Dataset.}\label{apex: add vigor}
\autoref{tab:Add Vigor} provides additional performance results on VIGOR dataset.  We provide this  as a reference for future studies in this domain. 

\subsection{Results on KITTI Test 2 Set with Angle Prior}\label{appex: kitti 2 prior}
\autoref{tab:prior test2} shows the results on KITTI test 2 set with angle prior of 10$^\circ$, 20$^\circ$ and 40$^\circ$. The results are reviewed in Section~\ref{section: with prior}.

\section{Memory and Time Performance on the KITTI Dataset}\label{sec: time memory petalview}
\autoref{tab: time memory petalview} shows the memory and time performance for PetalView on the KITTI dataset without location prior. Since the batch-wise early-stopping training strategy is applied, the search for each batch may be terminated before reaching the last level. Therefore, the training time is shorter in the early epochs.
\begin{table}[!h]
    \centering
    \begin{tabular}{|c|c|} \hline 
 \multicolumn{2}{|c|}{Training, batch size = 6}\\ \hline 
         Parameters&  $121.9 \times 10^6$\\ \hline 
         Speed early epoch&  5.8 samples/s\\ \hline 
         Time first epoch&  56 mins\\ \hline 
         Speed later epoch&  3.9 samples/s\\ \hline 
         Time later epoch&  85 mins\\ \hline 
 Memory GPU&40.2 Gb\\ \hline 
 \multicolumn{2}{|c|}{Testing, batch size = 16}\\ \hline 
 Speed&18.2 samples/s\\ \hline 
 Memory GPU&32.2 Gb\\ \hline
    \end{tabular}
    \caption[Memory and time performance on the KITTI Dataset in local-scale search]{Memory and time performance on KITTI Dataset in local-scale search. The training time is shorter in the first few epochs because the batch-wise early-stopping training strategy is applied.}
    \label{tab: time memory petalview}
\end{table}

\end{document}